\relax
\documentclass[11pt,a4paper]{article}
\usepackage[nohyperref]{naaclhlt2019}
\usepackage{times}
\usepackage{latexsym}

\usepackage{url}

\usepackage{amsthm}
\usepackage{amsmath}
\usepackage{amssymb}
\usepackage{algorithm}
\usepackage{algpseudocode}
\usepackage{algorithmicx}
\usepackage{hyperref}
\usepackage{booktabs}
\usepackage{makecell}
\usepackage{xcolor}
\usepackage{subcaption}
\usepackage{adjustbox}
\usepackage{multirow}

\newcommand{\comment}[1]{}  
\usepackage{todonotes} 
\tikzset{inlinenotestyle/.append style={align=justify}}

\newcommand{\lgspace}{\vspace{0mm}}
\newcommand{\midspace}{\vspace{0mm}}
\newcommand{\smspace}{\vspace{0mm}}

\aclfinalcopy 
\def\aclpaperid{772} 

\title{Multi-task Learning with Sample Re-weighting \\
for Machine Reading Comprehension}
\author{Yichong Xu$^1$\thanks{ \;\;Most of this work was performed when the author was interning at Microsoft.}, Xiaodong Liu$^{2}$, Yelong Shen$^{2}$, Jingjing Liu$^{2}$ and Jianfeng Gao$^{2}$ \\
	$^1$ Carnegie Mellon University\\
	$^2$ Microsoft Research\\
	\texttt{yichongx@cs.cmu.edu}\\
	\texttt{\{xiaodl, yeshen, jingjl,jfgao\}@microsoft.com }
}
\date{}
 
\begin{document}
\maketitle
\begin{abstract}

We propose a multi-task learning framework to learn a joint Machine Reading Comprehension (MRC) model that can be applied to a wide range of MRC tasks in different domains.  
Inspired by recent ideas of data selection in machine translation, we develop a novel sample re-weighting scheme to assign sample-specific weights to the loss.  
Empirical study shows that our approach can be applied to many existing MRC models. Combined with contextual representations from pre-trained language models (such as ELMo), we achieve new state-of-the-art results on a set of MRC benchmark datasets. We release our code at \url{https://github.com/xycforgithub/MultiTask-MRC}.
\end{abstract}
\midspace
\section{Introduction}
\smspace

Machine Reading Comprehension (MRC) has gained growing interest in the research community \cite{rajpurkar2016squad,yu2018qanet}. In an MRC task, the machine reads a text passage and a question, and generates (or selects) an answer based on the passage. This requires the machine to possess strong comprehension, inference and reasoning capabilities. Over the past few years, there has been much progress in building end-to-end neural network models \cite{seo2016bidirectional} for MRC. However, most public MRC datasets (e.g., SQuAD, MS MARCO, TriviaQA) are typically small (less than 100K) compared to the model size (such as SAN \cite{liu2017stochastic,liu2018sanv2} with around 10M parameters). To prevent over-fitting, recently there have been some studies on using pre-trained word embeddings \cite{pennington2014glove} and contextual embeddings in the MRC model training, as well as back-translation approaches \cite{yu2018qanet} for data augmentation. 

Multi-task learning \cite{caruana1997multitask} is a widely studied area in machine learning, aiming at better model generalization by combining training datasets from multiple tasks. In this work, we explore a multi-task learning (MTL) framework to enable the training of one universal model across different MRC tasks for better generalization. Intuitively, this multi-task MRC model can be viewed as an implicit data augmentation technique, which can improve generalization on the target task by leveraging training data from auxiliary tasks. 
We observe that merely adding more tasks cannot provide much improvement on the target task. Thus, we propose two MTL training algorithms to improve the performance.
The first method simply adopts a sampling scheme, which randomly selects training data from the auxiliary tasks controlled by a ratio hyperparameter;
The second algorithm incorporates recent ideas of data selection in machine translation \cite{van2017dynamic}. It learns the sample weights from the auxiliary tasks automatically through language models. 

Prior to this work, many studies have used upstream datasets to augment the performance of MRC models, including word embedding \cite{pennington2014glove}, language models (ELMo) \cite{peters2018deep} and machine translation \cite{yu2018qanet}. These methods aim to obtain a robust semantic encoding of both passages and questions. Our MTL method is orthogonal to these methods: rather than enriching semantic embedding with external knowledge, we leverage existing MRC datasets across different domains, which help make 
the whole comprehension process 
more robust and universal. Our experiments show that MTL can bring further performance boost when combined with contextual representations from pre-trained language models, e.g., ELMo \cite{peters2018deep}.

To the best of our knowledge, this is the first work that systematically explores multi-task learning for MRC. In previous methods that use language models and word embedding, the external embedding/language models are pre-trained separately and remain fixed during the training of the MRC model. Our model, on the other hand, can be trained with more flexibility on various MRC tasks. MTL is also faster and easier to train than embedding/LM methods: our approach requires no pre-trained models, whereas back translation and ELMo both rely on large models that would need days to train on multiple GPUs \cite{jozefowicz2016exploring,peters2018deep}.

We validate our MTL framework with two state-of-the-art models on four datasets from different domains.
Experiments show that our methods lead to a significant performance gain over single-task baselines on SQuAD \cite{rajpurkar2016squad}, NewsQA \cite{trischler2016newsqa} and Who-Did-What \cite{onishi2016did}, while achieving state-of-the-art performance on the latter two.
For example, on NewsQA \cite{trischler2016newsqa}, our model surpassed human performance by \textbf{13.4} (46.5 vs 59.9) and \textbf{3.2} (72.6 vs 69.4) absolute points in terms of exact match and F1.   

The contribution of this work is three-fold. First, we apply multi-task learning to the MRC task, which brings significant improvements over single-task baselines. Second, the performance gain from MTL can be easily combined with existing methods to obtain further performance gain. Third, the proposed sampling and re-weighting scheme can further improve the multi-task learning performance.
\midspace
\section{Related Work}
\smspace
Studies in machine reading comprehension mostly focus on architecture design of neural networks, such as bidirectional attention \cite{seo2016bidirectional}, dynamic reasoning \cite{xu2017dynamic}, and parallelization \cite{yu2018qanet}. Some recent work has explored transfer learning that leverages out-domain data to learn MRC models when no training data is available for the target domain \cite{golub2017two}. In this work, we explore multi-task learning to make use of the data from other domains, while we still have access to target domain training data.

Multi-task learning \cite{caruana1997multitask} has been widely used in machine learning to improve generalization using data from multiple tasks. 
For natural language processing, MTL has been successfully applied to low-level parsing tasks \cite{collobert2011natural}, sequence-to-sequence learning \cite{luong2015multi}, and web search \cite{liu2015representation}.
More recently, \cite{mccann2018natural} proposes to cast all tasks from parsing to translation as a QA problem and use a single network to solve all of them. However, their results show that multi-task learning hurts the performance of most tasks when tackling them together. Differently, we focus on applying MTL to the MRC task and show significant improvement over single-task baselines. 

Our sample re-weighting scheme bears some resemblance to previous MTL techniques that assign weights to tasks \cite{kendall2017multi}. However, our method gives a more granular score for each sample and provides better performance for multi-task learning MRC.

\midspace
\section{Model Architecture}
\smspace

We call our model Multi-Task-SAN (MT-SAN), which is a variation of SAN \cite{liu2017stochastic} model with two main differences: i) we add a highway network layer after the embedding layer, the encoding layer and the attention layer; 
ii) we use exponential moving average \cite{seo2016bidirectional} during evaluation. The SAN architecture and our modifications are briefly described below and in Section \ref{sec:expr_setup}, and detailed description can be found in \cite{liu2017stochastic}.
 
\midspace
\subsection{Input Format}
For most tasks we consider, our MRC model takes a triplet $(Q,P,A)$ as input, where $Q=(q_1,...,q_m), P=(p_1,...,p_n)$ are the word index representations of a question and a passage, respectively 
, and 
$A=(a_{\text{begin}}, a_{\text{end}})$ is the index of the answer span. 
The goal is to predict $A$ given $(Q,P)$. 

\midspace
\subsection{Lexicon Encoding Layer} 
We map the word indices of $P$ and $Q$ into their 300-dim Glove vectors \cite{pennington2014glove}. 
We also use the following additional information for embedding words: 
i) 16-dim part-of-speech (POS) tagging embedding; 
ii) 8-dim named-entity-recognition (NER) embedding; 
iii) 3-dim exact match embedding:
$f_{\text{exact\_match}}(p_i)=\mathbb{I}(p_i\in Q)$,
where matching is determined based on the original word, lower case, and lemma form, respectively; 
iv) Question enhanced passage word embeddings: $f_{\text{align}}(p_i)=\sum_{j} \gamma_{i,j} h(\text{GloVe}(q_j))$, where
\begin{align}
    \resizebox{0.89\hsize}{!}{
    $\gamma_{i,j}=\frac{\exp(h(\text{GloVe}(p_j)),h(\text{GloVe}(q_i)))}{\sum_{j'}\exp(h(\text{GloVe}(p_{j'})),h(\text{GloVe}(q_i)))}$}
\end{align}
is the similarity between word $p_j$ and $q_i$, and $g(\cdot)$ is a 300-dim single layer neural net with Rectified Linear Unit (ReLU) $g(x)=\text{ReLU}(W_1x)$; 
v) Passage-enhanced question word embeddings: the same as iv) but computed in the reverse direction. To reduce the dimension of the input to the next layer, the 624-dim input vectors of passages and questions are passed through a ReLu layer to reduce their dimensions to 125.

After the ReLU network, we pass the 125-dim vectors through a highway network \cite{srivastava2015highway}, to adapt to the multi-task setting:
$g_i = \text{sigmoid}(W_2p_i^t), p_i^t=\text{ReLU}(W_3p_i^t)\odot g_i + g_i\odot p_i^t$, where $p_i^t$ is the vector after ReLU transformation. Intuitively, the highway network here provides a neuron-wise weighting, which can potentially handle the large variation in data introduced by multiple datasets.

\midspace
\subsection{Contextual Encoding Layer} 
Both the passage and question encodings go through a 2-layer Bidirectional Long-Short Term Memory (BiLSTM, \citeauthor{hochreiter1997long}, \citeyear{hochreiter1997long}) network in this layer. We append a 600-dim CoVe vector \cite{mccann2017learned} to the output of the lexicon encoding layer as input to the contextual encoders. For the experiments with ELMo, we also append a 1024-dim ELMo vector. Similar to the lexicon encoding layer, the outputs of both layers are passed through a highway network for multi-tasking. Then we concatenate the output of the two layers to obtain $H^q\in \mathbb{R}^{2d\times m}$ for the question and $H^p=\mathbb{R}^{2d\times n}$ the passage, where $d$ is the dimension of the BiLSTM.

\midspace
\subsection{Memory/Cross Attention Layer}
We fuse $H^p$ and $H^q$ through cross attention and generate a working memory in this layer. We adopt the attention function from \cite{vaswani2017attention} and compute the attention matrix as
    \(C=\text{dropout}\left(f_{\text{attention}}(\hat{H}^q, \hat{H}^p)\right) \in\mathbb{R}^{m\times n}.\)
We then use $C$ to compute a question-aware passage representation as
$U^p = \text{concat}(H^p, H^qC)$. 
Since a passage usually includes several hundred tokens, we use the method of \cite{lin2017structured} to apply self attention to the representations of passage to rearrange its information:
   \( \hat{U}^p = U^p\text{drop}_{\text{diag}}(f_{\text{attention}}(U^p, U^p)),\)
where $\text{drop}_{\text{diag}}$ means that we only drop diagonal elements on the similarity matrix (i.e., attention with itself). Then, we concatenate $U^p$ and $\hat{U}^p$ and pass them through a BiLSTM: $M=\text{BiLSTM}([U^p];\hat{U}^p])$. Finally, output of the BiLSTM (after concatenating two directions) goes through a highway layer to produce the memory.

\midspace
\subsection{Answer Module}
The base answer module is the same as SAN, which computes a distribution over spans in the passage.
Firstly, 
we compute an initial state $s_0$ by self attention on $H^q$: $s_0\leftarrow \text{Highway}\left(\sum_{j} \frac{\exp(w_4H^q_j)}{\sum_{j'}\exp{w_4H^q_{j'}}}\cdot H^q_j\right)$. The final answer is computed through $T$ time steps. At step $t\in \{1,...,T-1\}$, we compute the new state using a Gated Recurrent Unit (GRU, \citeauthor{cho2014properties}, \citeyear{cho2014properties}) $s_t=\text{GRU}(s_{t-1},x_t)$, where $x_t$ is computed by attention between $M$ and $s_{t-1}$: $x_t=\sum_{j} \beta_j M_j, \beta_j=\text{softmax}(s_{t-1}W_5M)$. Then each step produces a prediction of the start and end of answer spans through a bilinear function:
\(P_t^{\text{begin}}=\text{softmax}(s_tW_6M), \)
\(P_t^{\text{end}}=\text{softmax}(s_tW_7M). \)
The final prediction is the average of each time step: $P^{\text{begin}}=\frac{1}{T}\sum_{t}P_t^{\text{begin}},P^{\text{end}}=\frac{1}{T}\sum_{t}P_t^{\text{end}}$. We randomly apply dropout on the step level in each time step during training, as done in \cite{liu2017stochastic}. During training, the objective is the log-likelihood of the ground truth: $l(Q,P,A)=\log P^{\text{begin}}(a_{\text{begin}})+\log P^{\text{end}}(a_{\text{end}})$. 






\midspace
\section{\label{sec:mtl_algo} Multi-task Learning Algorithms}
\smspace

We describe our MTL training algorithms in this section.
We start with a very simple and straightforward algorithm that samples one task and one mini-batch from that task at each iteration. To improve the performance of MTL on a target dataset, we propose two methods to re-weight samples according to their importance. The first proposed method directly lowers the probability of sampling from a particular auxiliary task; however, this probability has to be chosen using grid search. We then propose another method that avoids such search by using a language model. 


\begin{algorithm}[h!]
	\caption{Multi-task Learning of MRC}
	\begin{algorithmic}[1]		
		\Require{k different datasets $\mathcal{D}_1,...,\mathcal{D}_K$, max\_epoch}
		\State Initialize the model $\mathcal{M}$
		\For{epoch$=1,2,...$, max\_epoch}
		\State Divide each dataset $\mathcal{D}_k$ into $N_k$ mini-batches $\mathcal{D}_k=\{b_1^k,...,b_{N_k}^k\}$, $1\leq k\leq K$ 
		\State Put all mini-batches together and randomly shuffle the order of them, to obtain a sequence $B=(b_1,...,b_L)$, where $L=\sum_k N_k$
		\For{each mini-batch $b\in B$}
		\State Perform gradient update on $\mathcal{M}$ with loss $l(b)=\sum_{(Q,P,A)\in b} l(Q,P,A)$ \label{step:loss}
		\EndFor
		\State Evaluate development set performance
		\EndFor
		\Ensure{Model with best evaluation performance 
		}
	\end{algorithmic}
	\label{algo:mtmrc}
	\smspace
\end{algorithm}


Suppose we have $K$ different tasks, the simplest version of our MTL training procedure is shown in Algorithm \ref{algo:mtmrc}. In each epoch, we take all the mini-batches from all datasets and shuffle them for model training, and the same set of parameters is used for all tasks. 
Perhaps surprisingly, as we will show in the experiment results, this simple baseline method can already lead to a considerable improvement over the single-task baselines. 

\midspace
\subsection{\label{sec:mtl_ratio} Mixture Ratio} 

One observation is that the performance of our model using Algorithm \ref{algo:mtmrc} starts to deteriorate as we add more and more data from other tasks into our training pool. We hypothesize that the external data will inevitably bias the model towards auxiliary tasks instead of the target task.



\begin{algorithm}[h!]
	\caption{Multi-task Learning of MRC with mixture ratio, targeting $\mathcal{D}_1$}
	\begin{algorithmic}[1]		
		\Require{K different datasets $\mathcal{D}_1,...,\mathcal{D}_K$, max\_epoch, mixture ratio $\alpha$}
		\State Initialize the model $\mathcal{M}$
		\For{epoch$=1,2,...$, max\_epoch}
		\State Divide each dataset $\mathcal{D}_k$ into $N_k$ mini-batches $\mathcal{D}_k=\{b_1^k,...,b_{N_k}^k\}$, $1\leq k\leq K$ 
		\State $S\leftarrow \{b_1^1,...,b_{N_1}^1\}$
		\State Randomly pick $\lfloor\alpha N_1 \rfloor$ mini-batches from $\bigcup_{k=2}^{K} \mathcal{D}_k$ and add to $S$
		\State Assign mini-batches in $S$ in a random order to obtain a sequence $B=(b_1,...,b_L)$, where $L=N_1+\lfloor\alpha N_1\rfloor$
		\For{each mini-batch $b\in B$}
		\State Perform gradient update on $\mathcal{M}$ with loss $l(b)=\sum_{(Q,P,A)\in b} l(Q,P,A)$
		\EndFor
		\State Evaluate development set performance
		\EndFor
		\Ensure{Model with best evaluation performance 
		}
	\end{algorithmic}
	\label{algo:mtmrc_ratio}
	\smspace
\end{algorithm}

To avoid such adverse effect, we introduce a mixture ratio parameter during training. The training algorithm with the mixture ratio is presented in Algorithm \ref{algo:mtmrc_ratio}, with $\mathcal{D}_1$ being the target dataset. In each epoch, we use all mini-batches from $\mathcal{D}_1$, while only a ratio $\alpha$ of mini-batches from external datasets are used to train the model. In our experiment, we use hyperparameter search to find the best $\alpha$ for each dataset combination. This method resembles previous methods in multi-task learning to weight losses differently (e.g., \citeauthor{kendall2017multi}, \citeyear{kendall2017multi}), and is very easy to implement. In our experiments, we use Algorithm \ref{algo:mtmrc_ratio} to train our network when we only use 2 datasets for MTL. 

\begin{table*}[htb!]
	\begin{center}
		\begin{tabular}{|c|c|c|c|c|}
			\hline \bf Dataset & SQuAD(v1) & NewsQA & MS MARCO(v1) & WDW \\ \hline
			\# Training Questions & 87,599  & 92,549 & 78,905 & 127,786 \\
			Text Domain & Wikipedia & CNN News & Web Search & Gigaword Corpus\\
			Avg. Document Tokens &  130 & 638 & 71 & 365\\
			Answer type & Text span & Text span & Natural sentence & Cloze\\
			Avg. Answer Tokens & 3.5 & 4.5 & 16.4 & N/A\\
			\hline
		\end{tabular}
	\end{center}
	\lgspace
	\caption{Statistics of the datasets. Some numbers come from
		\protect\cite{sugawara2017evaluation}.
	}
	\label{tab:datasets}
\lgspace
\end{table*}

\midspace
\subsection{\label{sec:mtl_weighting} Sample Re-Weighting}
The mixture ratio (Algorithm~\ref{algo:mtmrc_ratio}) dramatically improves the performance of our system. However, it requires to find an ideal ratio by hyperparameter search which is time-consuming. Furthermore, 
the ratio gives the same weight to every auxiliary data, but the relevance of every data point to the target task can vary greatly.

We develop a novel re-weighting method to resolve these problems, using ideas inspired by data selection in machine translation \cite{axelrod2011domain,van2017dynamic}. We use 
$(Q^{k},P^{k},A^{k})$ to represent a data point from the $k$-th task for $1\leq k\leq K$, with $k=1$ being the target task. Since the passage styles are hard to evaluate, we only evaluate data points based on $Q^{k}$ and $A^k$. Note that only data from auxiliary task ($2\leq k\leq K$) is re-weighted; target task data always have weight 1.

Our scores consist of two parts, one for questions and one for answers. For questions, we create language models (detailed in Section \ref{sec:expr_setup}) using questions from each task, which we represent as $LM_k$ for the $k$-th task. For each question $Q^{k}$ from auxiliary tasks, we compute a cross-entropy score:
\begin{align}
    H_{C,Q}(Q^{k})=-\frac{1}{m}\sum_{w\in Q^{k}}\log (LM_{C}(w)),
\end{align}
where $C\in \{1,k\}$ is the target or auxiliary task,  $m$ is the length of question $Q^{k}$, and $w$ iterates over all words in $Q^{k}$. 

It is hard to build language models for answers since they are typically very short (e.g., answers on SQuAD includes only one or two words in most cases). 
We instead just use the length of answers as a signal for scores. Let $l_{a}^{k}$ be the length of $A^{k}$, the cross-entropy answer score is defined as:
\begin{align}
    H_{C,A}(A^{k})=-\log \text{freq}_C(l_a^{k}),
\end{align}
where freq$_C$ is the frequency of answer lengths in task $C\in \{1,k\}$. 

The cross entropy scores are then normalized over all samples in task $C$ to create a comparable metric across all auxiliary tasks:
\begin{align}
    H_{C,Q}'(Q^k)=\frac{H_{C,Q}(Q^k)-\min(H_{C,Q})}{\max(H_{C,Q})-\min(H_{C,Q})} \\
    H_{C,A}'(A^k)=\frac{H_{C,A}(A^k)-\min(H_{C,A})}{\max(H_{C,A})-\min(H_{C,A})} 
\end{align}
for $C\in \{1,2,...,K\}$. 
For $C\in \{2,...,K\}$, the maximum and minimum are taken over all samples in task $k$. For $C=1$ (target task), they are taken over all available samples.

Intuitively, $H'_{C,Q}$ and $H'_{C,A}$ represents the similarity of text $Q,A$ to task $C$; a low $H'_{C,Q}$ (resp. $H'_{C,A}$) means that $Q^k$ (resp. $A^k$) is easy to predict and similar to $C$, and vice versa. We would like samples that are most similar from data in the target domain (low $H'_1$), and most different (informative) from data in the auxiliary task (high $H'_{k}$). We thus compute the following cross-entropy difference for each external data:
\begin{align}
\text{CED}(Q^{k},A^{k})=&(H'_{1,Q}(Q^{k})-H'_{k,Q}(Q^{k}))+\nonumber\\
&(H'_{1,A}(A^{k})-H'_{k,A}(A^{k})) \label{eqn:ced}
\end{align}
for $k\in \{2,...,K\}$. Note that a low CED score indicates high importance. 
Finally, we transform the scores to weights by taking negative, and normalize between $[0,1]$:
\begin{align}
\resizebox{0.89\hsize}{!}{$\displaystyle \text{CED}'(Q^{k},A^{k})
=1-\frac{\text{CED}(Q^{k},A^{k})-\min(\text{CED})}{\max(\text{CED})-\min(\text{CED})}.\label{eqn:cedp}$}
\end{align} 

Here the maximum and minimum are taken over all available samples and task. Our training algorithm is the same as Algorithm 1, but for minibatch $b$ we instead use the loss 
\begin{align}
    l(b)=\sum_{(P,Q,A)\in b} \text{CED}'(Q,A)l(P,Q,A)
\end{align}
in step \ref{step:loss}. We define $\text{CED}'(Q^1,A^1)\equiv 1$ for all target samples $(P^1,Q^1,A^1)$.

\begin{table*}[htb!]
	\begin{center}
		\begin{tabular}{l c}
			\hline  Model & Dev Set Performance   \\ \toprule
			Single Model without Language Models & EM,F1 \\   \midrule
			BiDAF \cite{seo2016bidirectional} & 67.7, 77.3 \\
			SAN \cite{liu2017stochastic} &76.24, 84.06 \\
			MT-SAN on SQuAD (single task, ours) & 76.84, 84.54 	\\
			MT-SAN on SQuAD+NewsQA(ours) &  78.60, 85.87 \\	
			MT-SAN on SQuAD+MARCO(ours) & 77.79, 85.23 \\	
			\bf MT-SAN on SQuAD+NewsQA+MARCO(ours) & \bf 78.72, 86.10 \\	\midrule
			Single Model with ELMo &  \\   \midrule
			SLQA+ \cite{wang2018multib} & 80.0, 87.0  \\
			MT-SAN on SQuAD (single task, ours) & 80.04, 86.54 	\\
			MT-SAN on SQuAD+NewsQA(ours) & 81.36, 87.71 \\	
			MT-SAN on SQuAD+MARCO(ours) & 80.37, 87.17 \\	
			\bf MT-SAN on SQuAD+NewsQA+MARCO(ours) & \bf 81.58, 88.19 \\	
			\bf BERT \cite{devlin2018bert} & \bf 84.2, 91.1 \\	
			\hline
			Human Performance (test set) & 82.30, 91.22 
		\end{tabular}
	\end{center}
	\lgspace
	\caption{\label{tab:mt-san-squad} Performance of our method to train SAN in multi-task setting, competing published results, leaderboard results and human performance, on SQuAD dataset (single model). Note that BERT uses a much larger language model, and is not directly comparable with our results. We expect our test performance is roughly similar or a bit higher than our dev performance, as is the case with other competing models.
	}
	\lgspace
\end{table*}

\midspace
\section{Experiments}
\smspace

Our experiments are designed to answer the following questions on multi-task learning for MRC:\\
1. Can we improve the performance of existing MRC systems using multi-task learning?\\
2. How does multi-task learning affect the performance if we combine it with other external data?\\
3. How does the learning algorithm change the performance of multi-task MRC?\\
4. How does our method compare with existing MTL methods?\\
We first present our experiment details and results for MT-SAN. Then, we provide a comprehensive study on the effectiveness of various MTL algorithms in Section \ref{sec:expr_mtl_algo}. 
At last, we provide some additional results on combining MTL with DrQA \cite{chen2017reading} to show the flexibility of our approach \footnote{We include the results in the appendix due to space limitations.}.

\midspace
\subsection{Datasets}

We conducted experiments on SQuAD (\citeauthor{rajpurkar2016squad}, \citeyear{rajpurkar2016squad}), NewsQA\cite{trischler2016newsqa}, MS MARCO (v1, \citeauthor{nguyen2016ms},\citeyear{nguyen2016ms}) and WDW \cite{onishi2016did}. Dataset statistics is shown in Table \ref{tab:datasets}. Although similar in size, these datasets are quite different in domains, lengths of text, and types of task. In the following experiments, we will validate whether including external datasets as additional input information (e.g., pre-trained language model on these datasets) helps boost the performance of MRC systems.

\midspace
\subsection{\label{sec:expr_setup} Experiment Details}
We mostly focus on span-based datasets for MT-SAN, namely SQuAD, NewsQA, and MS MARCO. 
We convert MS MARCO into an answer-span dataset to be consistent with SQuAD and NewsQA, following \cite{liu2017stochastic}. For each question, we search for the best span using ROUGE-L score in all passage texts and use the span to train our model. We exclude questions with maximal ROUGE-L score less than 0.5 during training. For evaluation, we use our model to find a span in all passages. The prediction score is multiplied with the ranking score, trained following \citet{liu2018stochastic}'s method to determine the final answer. 

We train our networks using algorithms in Section \ref{sec:mtl_algo}, using SQuAD as the target task. For experiments with two datasets, we use Algorithm \ref{algo:mtmrc_ratio}; for experiments with three datasets we find the re-weighting mechanism in Section \ref{sec:mtl_weighting} to have a better performance (a detailed comparison will be presented in Section \ref{sec:expr_mtl_algo}). 

For generating sample weights, we build a LSTM language model on questions following the implementation of \citet{merityRegOpt} with the same hyperparameters. We only keep the 10,000 most frequent words, and replace the other words with a special out-of-vocabulary token. 

Parameters of MT-SAN are mostly the same as in the original paper \cite{liu2017stochastic}. We utilize spaCy\footnote{\href{https://spacy.io}{https://spacy.io}} to tokenize the text and generate part-of-speech and named entity labels. We use a 2-layer BiLSTM with 125 hidden units as the BiLSTM throughout the model. During training, we drop the activation of each neuron with 0.3 probability. 
For optimization, we use Adamax \cite{kingma2014adam} with a batch size of 32 and a learning rate of 0.002. For prediction, we compute an exponential moving average (EMA, \citeauthor{seo2016bidirectional} \citeyear{seo2016bidirectional}) of model parameters with a decay rate of 0.995 and use it to compute the model performance. 
 For experiments with ELMo, we use the model implemented by AllenNLP \footnote{\href{https://allennlp.org/}{https://allennlp.org/}}. 
We truncate passage to contain at most 1000 tokens during training and eliminate those data with answers located after the 1000th token. 
The training converges in around 50 epochs for models without ELMo (similar to the single-task SAN); For models with ELMo, the convergence is much faster (around 30 epochs). 

\midspace
\subsection{Performance of MT-SAN}
\label{sub:result-mt-san}
In the following sub-sections, we report our results on SQuAD and MARCO development sets, as well as on the development and test sets of NewsQA \footnote{
 The official submission for SQuAD v1.1 and MARCO v1.1 are closed, so we report results on the development set. According to their leaderboards, performances on development and test sets are usually similar.}. All results are single-model performance unless otherwise noted.

The multi-task learning results of SAN on SQuAD are summarized in Table \ref{tab:mt-san-squad}. By using MTL on SQuAD and NewsQA, we can improve the exact-match (EM) and F1 score by (2\%, 1.5\%), respectively, both with and without ELMo. The similar gain indicates that our method is orthogonal to ELMo. Note that our single-model performance is slightly higher than the original SAN, by incorporating EMA and highway networks. By incorporating with multi-task learning, it further improves the performance. The performance gain by adding MARCO is relatively smaller, with 1\% in EM and 0.5\% in F1. We conjecture that MARCO is less helpful due to its differences in both the question and answer style. For example, questions in MS MARCO are real web search queries, which are short and may have typos or abbreviations; while questions in SQuAD and NewsQA are more formal and well written.  

Using 3 datasets altogether provides another marginal improvement.
Our model obtains the best results among existing methods that do not use a large language model (e.g., ELMo). Our ELMo version also outperforms any other models which are under the same setting. We note that BERT \cite{devlin2018bert} uses a much larger model than ours(around 20x), and we leave the performance of combining BERT with MTL as interesting future work.
\begin{table}[htb!]
	\begin{center}
		\begin{tabular}{l c c}
			\hline  Model & Dev Set & Test Set   \\ \hline
			Model W/o ELMo & EM,F1 & EM, F1\\   \midrule
			Match-LSTM$^1$  & 34.4, 49.6 & 34.9, 50.0\\
			FastQA$^2$ & 43.7, 56.1 & 42.8, 56.1 \\
			AMANDA$^3$ & 48.4, 63.3 & 48.4, 63.7 \\
			MT-SAN (Single task)  & 55.8, 67.9 & 55.6, 68.0 \\
			\bf MT-SAN (S+N) & \bf 57.8, 69.9 & \bf 58.3, 70.7 \\	
			\hline
			Model With ELMo & &\\\hline
			MT-SAN (Single task) & 57.7, 70.4 & 57.0, 70.4 \\
			\bf MT-SAN (S+N) & \bf 60.1, 72.5 & \bf 59.9, 72.6 \\\hline	
			Human Performance & -,- & 46.5, 69.4\\
			\hline
		\end{tabular}
	\end{center}
	\lgspace
	\caption{\label{tab:mt-san-newsqa} Performance of our method to train SAN in multi-task setting, with published results and human performance on NewsQA dataset. All SAN results are from our models. ``S+N'' means jointly training on SQuAD and NewsQA
	References: $^1$: implemented by \protect\citeauthor{trischler2016newsqa} (\citeyear{trischler2016newsqa}). 
    $^2$:\protect\citeauthor{weissenborn2017making}(\citeyear{weissenborn2017making}).
    $^3$: \protect\citeauthor{kundu2018question}(\citeyear{kundu2018question}).
}
\lgspace
\end{table}

The results of multi-task learning on NewsQA are in Table \ref{tab:mt-san-newsqa}. The performance gain with multi-task learning is even larger on NewsQA, with over 2\% in both EM and F1. Experiments with and without ELMo give similar results. What is worth noting is that our approach not only achieves new state-of-art results with a large margin but also surpasses human performance on NewsQA. 

Finally we report MT-SAN performance on MS MARCO in Table \ref{tab:mt-san-marco}. Multi-tasking on SQuAD and NewsQA provides a similar performance boost in terms of BLEU-1 and ROUGE-L score as in the case of NewsQA and SQuAD. Our method does not achieve very high performance compared to previous work, probably because we do not apply common techniques like yes/no classification or cross-passage ranking \cite{wang2018multi}.

\begin{table}[tb!]
	\begin{center}
		\begin{tabular}{l c}
			\hline  Model & Scores   \\ \hline
			Single Model W/o ELMo &  \\   \midrule
			FastQAExt$^1$ (test set)  & 33.99, 32.09\\
			Reasonet++$^2$ & 38.62, 38.01 	\\
			V-Net$^3$ & -, 45.65\\
			SAN$^4$ & 43.85, 46.14\\
			MT-SAN  & 34.13, 42.65 \\
MT-SAN: SQuAD+MARCO & 34.29, 43.47  \\	
\bf MT-SAN: 3 datasets & \bf 36.99, 43.64\\
\hline
Single Model With ELMo & \\\hline
MT-SAN & 34.57, 42.88 \\
MT-SAN: SQuAD+MARCO & 37.02, 43.89 \\	
\bf MT-SAN: 3 datasets & \bf 37.12, 44.12\\	\hline	
			Human Performance (test set) & 48.02, 49.72\\
		\end{tabular}
	\end{center}
	\caption{\label{tab:mt-san-marco} Performance of our method to train SAN in multi-task setting, competing published results and human performance, on MS MARCO dataset. The scores stand for (BLEU-1, ROUGE-L) respectively. All SAN results are our results. ``3 dataset'' means we train using SQuAD+NewsQA+MARCO. References: $^1$: \protect\cite{weissenborn2017making}. $^2$: implemented by \protect\cite{shen2017empirical}. $^3$:\protect\cite{wang2018multi}. $^4$: \protect\cite{liu2017stochastic}}
\end{table}


\begin{table}[htb!]
	\begin{center}
		\begin{tabular}{l  c  c}
			\hline  Model &  SQuAD & WDW  \\ \hline
			MT-SAN (Single Task) & 76.8, 84.5 & 77.5 \\ 
            MT-SAN (S+W) & \bf 77.6, 85.1 &\bf 78.5 \\ 
            \hline
            SOTA\cite{yang2016words}. & 86.2, 92.2& 71.7\\
            Human Performance & 82.3, 91.2 & 84\\\hline 
		\end{tabular}
	\end{center}
	\lgspace
	\caption{\label{tab:wdw} Performance of MT-SAN on SQuAD Dev and WDW test set. Accuracy is used to evaluate WDW. ``S+W'' means jointly training on SQuAD and WDW.
    }
    \lgspace
\end{table}

We also test the robustness of our algorithm by performing another set of experiments on SQuAD and WDW. WDW is much more different than the other three datasets (SQuAD, NewsQA, MS MARCO): WDW guarantees that the answer is always a person, whereas the percentage of such questions in SQuAD is 12.9\%. Moreover, WDW is a cloze dataset, whereas in SQuAD and NewsQA answers are spans in the passage. We use a task-specific answer layer in this experiment and use Algorithm \ref{algo:mtmrc_ratio}; the WDW answer module is the same as in AS Reader \cite{kadlec2016text}, which we describe in the appendix for completeness. 
Despite these large difference between datasets, our results (Table \ref{tab:wdw}) show that MTL can still provide a moderate performance boost when jointly training on SQuAD (around 0.7\%) and WDW (around 1\%).

\begin{table}[tb!]
	\begin{center}
		\begin{tabular}{l  c  c}
			\hline  Model &  EM, F1 & +/-  \\ \hline
			QANet & 73.6, 82.7 & 0.0, 0.0\\
            QANet + BT & 75.1, 83.8 & +1.5,+1.1\\ \hline
            SAN & 76.8, 84.5 & 0.0, 0.0 \\
            \bf MT-SAN & \bf 78.7, 86.0 &\bf +1.9,+1.5 \\
            SAN + ELMo & 80.0, 86.5 & +3.2,+2.0\\
            \bf MT-SAN + ELMo&\bf 81.6, 88.2 &\bf +4.8, +3.7\\
            \hline
		\end{tabular}
	\end{center}
	\lgspace
	\caption{\label{tab:comp-augment} Comparison of methods to use external data. BT stands for back translation \cite{yu2018qanet}.
    }
    \lgspace
\end{table}
\noindent\textbf{Comparison of methods using external data.} As a method of data augmentation, we compare our approach to previous methods for MRC in Table \ref{tab:comp-augment}. Our model achieves better performance than back translation. We also observe that language models such as ELMo obtain a higher performance gain than multi-task learning, however, combining it with multi-task learning leads to the most significant performance gain. This validates our assumption that multi-task learning is more robust and is different from previous methods such as language modeling.

\begin{table}[tb!]
	\begin{center}
		\begin{tabular}{@{\hskip1pt}l@{\hskip1pt} c @{\hskip1pt} }
			\hline Model & Performance \\
			\hline  SQuAD + MARCO& EM,F1 \\ \hline
			Simple Combine (Alg. \ref{algo:mtmrc}) & 77.1, 84.6\\
			Loss Uncertainty \cite{kendall2017multi} & 77.3, 84.7 \\
			Mixture Ratio & 77.8, 85.2 \\
			\bf Sample Re-weighting & \bf 77.9,85.3\\
			\hline  SQuAD + NewsQA + MARCO&  \\ \hline
			Simple Combine (Alg. \ref{algo:mtmrc}) & 77.6, 85.2\\
			Loss Uncertainty \cite{kendall2017multi} & 78.2, 85.6\\
			Mixture Ratio & 78.4, 85.7 \\
			\textbf{Sample Re-weighting} & \textbf{78.8}, \textbf{86.0}\\\hline			
		\end{tabular}
	\end{center}
	\lgspace
	\caption{\label{tab:mtl_algo} Comparison of different MTL strategies on MT-SAN. Performance is on SQuAD.
	}
	\lgspace
\end{table}

\begin{table*}[tb!]
	\begin{center}
		\begin{tabular}{c|l|c|c|c}
			\hline & \bf Samples/Groups & CED$'$ & $H_Q$ & $H_A$  \\ \hline
			\multirow{4}{*}{Examples}& (NewsQA) Q: Where is the drought hitting? & \multirow{2}{*}{0.824} & \multirow{2}{*}{0.732} & \multirow{2}{*}{0.951}\\
& A: Argentina &&&\\\cline{2-5}
&(MARCO) Q: thoracic cavity definition   & \multirow{2}{*}{0.265} & \multirow{2}{*}{0.332} & \multirow{2}{*}{0.240}\\
&A: is the chamber of the human body ... and fascia.&&&\\\hline
			\multirow{4}{*}{Averages}&Samples in NewsQA & 0.710 & 0.593 & 0.895\\
			&Samples in MARCO & 0.587 & 0.550& 0.669\\
			&MARCO Questions that start with ``When'' or ``Who'' & 0.662  & 0.605 & 0.761\\
			&All samples & 0.654& 0.573 & 0.791\\\hline
		\end{tabular}
	\end{center}
	\lgspace
	\caption{\label{tab:scores} Scores for examples from NewsQA and MS MARCO and average scores for specific groups of samples. CED$'$ is as in (\ref{eqn:cedp}), while $H_Q$ and $H_A$ are normalized version of question and sample scores.
	``Sum'' are the actual scores we use, and ``LM'', ``Answer'' are scores from language models and answer lengths. }
	\lgspace
\end{table*}

\midspace
\subsection{\label{sec:expr_mtl_algo} Comparison of Different MTL Algorithms}
In this section, we provide ablation studies as well as comparisons with other existing algorithms on the MTL strategy. We focus on MT-SAN without ELMo for efficient training. 

Table \ref{tab:mtl_algo} compares different multi-task learning strategies for MRC. 
Both the mixture ratio (Sec \ref{sec:mtl_ratio}) and sample re-weighting (Sec \ref{sec:mtl_weighting}) improves over the naive baseline of simply combining all the data (Algorithm \ref{algo:mtmrc}). On SQuAD+MARCO, they provide around 0.6\% performance boost in terms of both EM and F1, and around 1\% on all 3 datasets. 
We note that this accounts for around a half of our overall improvement. Although sample re-weighting performs similar as mixture ratio, it significantly reduces the amount of training time as it eliminates the need for a grid searching the best ratio. 
Kendal et al., (\citeyear{kendall2017multi}) use task uncertainty to weight tasks differently for MTL; our experiments show that this has some positive effect, but does not perform as well as our proposed two techniques. 
We note that Kendal et al. (as well as other previous MTL methods) optimizes the network to perform well for all the tasks, whereas our method focuses on the target domain which we are interested in, e.g., SQuAD. 



\begin{figure}[tb!]
	\centering
	\includegraphics[width=0.45\textwidth]{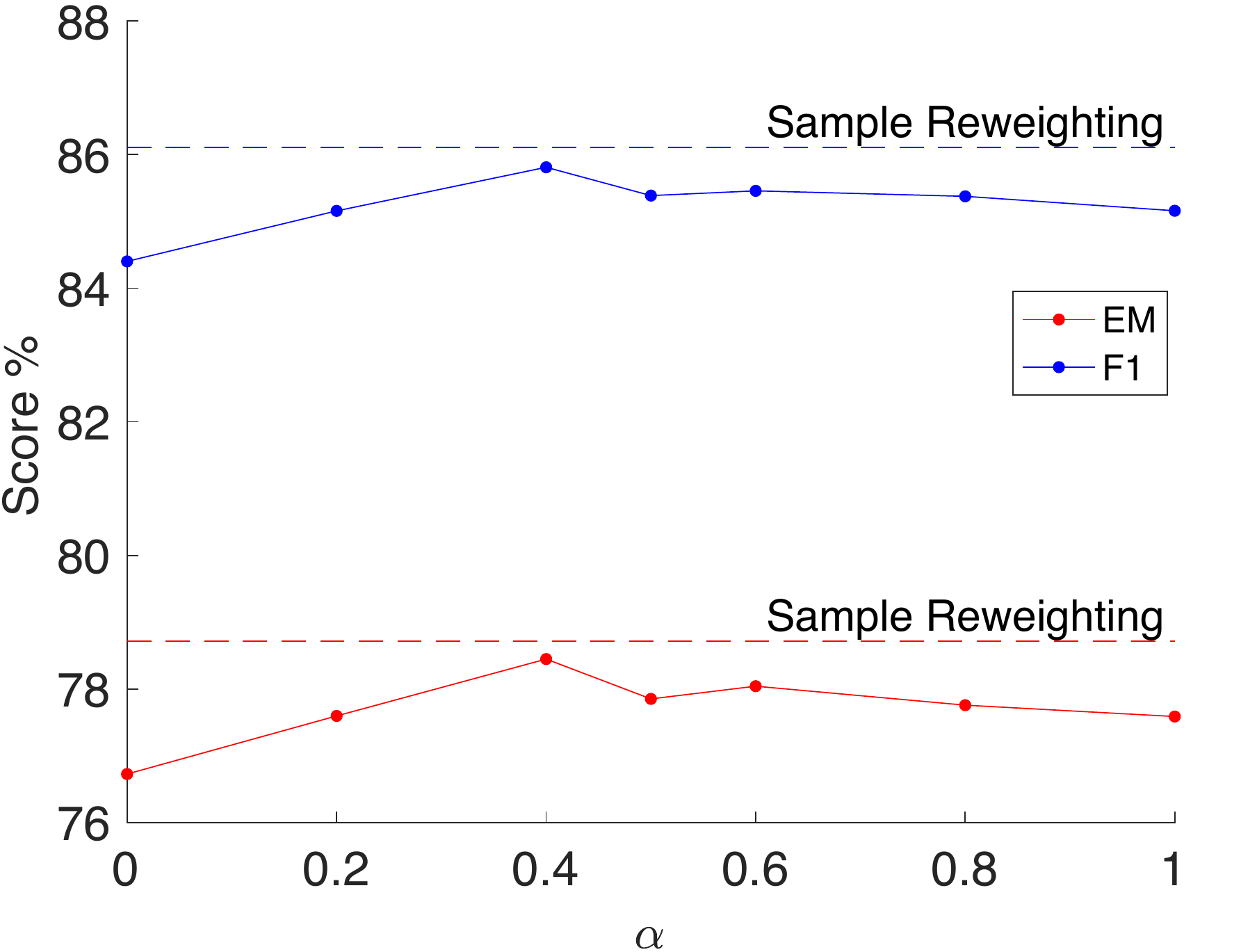}
	\lgspace
	\caption{\label{fig:ratio_snm} Effect of the mixture ratio on the performance of MT-SAN. Note that $\alpha=0$ is equivalent to single task learning, and $\alpha=1$ is equivalent to simple combining.}
	\lgspace
\end{figure}
\noindent\textbf{Sensitivity of mixture ratio.} We also investigate the effect of mixture ratio on the model performance. We plot the EM/F1 score on SQuAD dev set vs. mixture ratio in Figure \ref{fig:ratio_snm} for MT-SAN when trained on all three datasets. The curve peaks at $\alpha=0.4$; however if we use $\alpha=0.2$ or $\alpha=0.5$, the performance drops by around $0.5\%$, well behind the performance of sample re-weighting. This shows that the performance of MT-SAN is sensitive to changes in $\alpha$, making the hyperparameter search even more difficult. Such sensitivity suggests a preference for using our sample re-weighting technique. On the other hand, the ratio based approach is pretty straightforward to implement. 

\noindent\textbf{Analysis of sample weights.} Dataset comparisons in Table \ref{tab:datasets} and performance in Table \ref{tab:mt-san-squad} suggests that NewsQA share more similarity with SQuAD than MARCO. Therefore, a MTL system should weight NewsQA samples more than MARCO samples for higher performance. We try to verify this in Table \ref{tab:scores} by showing examples and statistics of the sample weights. We present the CED$'$ scores, as well as normalized version of question and answer scores (resp. $(H'_{1,Q}-H'_{k,Q})$ and $(H'_{1,A}-H'_{k,A})$ in (\ref{eqn:ced}), and then negated and normalized over all samples in NewsQA and MARCO in the same way as in (\ref{eqn:cedp})). A high $H_Q$ score indicates high importance of the question, and $H_A$ of the answer; CED$'$ is a summary of the two.
We first show one example from NewsQA and one from MARCO. The NewsQA question is a natural question (similar to SQuAD) with a short answer, leading to high scores both in questions and answers. The MARCO question is a phrase, with a very long answer, leading to lower scores. From overall statistics, we also find samples in NewsQA have a higher score than those in MARCO. However, if we look at MARCO questions that start with ``when'' or ``who'' (i.e., probability natural questions with short answers), the scores go up dramatically.

\lgspace
\section{Conclusion}
\midspace
We proposed a multi-task learning framework to train MRC systems using datasets from different domains and developed two approaches to re-weight the samples for multi-task learning on MRC tasks. Empirical results demonstrated our approaches outperform existing MTL methods and the single-task baselines as well. Interesting future directions include combining with larger language models such as BERT, and MTL with broader tasks such as language inference \cite{liu2019multi} and machine translation.

\section*{Acknowledgements}	
Yichong Xu has been partially supported by DARPA (FA8750-17-2-0130).

\bibliography{yichongref}
\bibliographystyle{acl_natbib}
\appendix
\onecolumn


\section{Answer Module for WDW}

We describe the answer module for WDW here for completeness. For WDW we need to choose an answer from a list of candidates; the candidates are people names that have appeared in the passage. We use the same way to summary information in questions as in span-based models: $s_0\leftarrow \text{Highway}\left(\sum_{j} \frac{\exp(w_4H^q_j)}{\sum_{j'}\exp{w_4H^q_{j'}}}\cdot H^q_j\right)$. We then compute an attention score via simple dot product: $s=\text{softmax}(s_0^TM)$. The probability of a candidate being the true answer is the aggregation of attention scores for all appearances of the candidate: 
$$\Pr(c|Q,P) \propto \sum_{1\leq i\leq n} s_i\mathbb{I}(p_i\in C) $$
for each candidate $C$. Recall that $n$ is the length of passage $P$, and $p_i$ is the i-th word; therefore $\mathbb{I}(p_i\in C)$ is the indicator function of $p_i$ appears in candidate $C$. The candidate with the largest probability is chosen as the predicted answer.

\section{Experiment Results on DrQA}
To demonstrate the flexibility of our approach, we also adapt DrQA \cite{chen2017reading} into our MTL framework. We only test DrQA using the basic Algorithm \ref{algo:mtmrc_ratio}, since our goal is mainly to test the MTL framework.
\subsection{Model Architecture}
Similar to MT-SAN, we add a highway network after the lexicon encoding layer and the contextual encoding layer and use a different answer module for each dataset. We apply MT-DrQA to a broader range of datasets. For span-detection datasets such as SQuAD, we use the same answer module as DrQA. For cloze-style datasets like Who-Did-What, we use the attention-sum reader \cite{kadlec2016text} as the answer module. For classification tasks required by SQuAD v2.0 \cite{rajpurkar2018know}, we apply a softmax to the last state in the memory layer and use it as the prediction. 
\subsection{Performance of MT-DrQA}
\begin{table*}[htb!]
	\begin{center}
		\begin{tabular}{l|c|c|c|c}
			\hline \bf Setup & SQuAD (v1) & SQuAD (v2) & NewsQA & WDW  \\ \hline
			Single Dataset & 69.5,78.8 (paper) & 61.9, 65.2 & 51.9, 64.6   & \bf 75.8 \\
            & 68.6, 77.8 \comment{68.63, 77.80} (ours) & & &\\
			MT-DrQA on Sv1+NA & 70.2, 79.3 \comment{70.21, 79.23} & -,- & 52.8, 65.8  & -\\
			MT-DrQA on Sv1+W & 69.2, 78.4 &-,-& -,- & 75.7 \\
			MT-DrQA on Sv1+N+W & \bf 70.2, 79.3 & -,- & \bf 53.1, 65.7  & 75.4 \\
			MT-DrQA on Sv2+N & -,- & \bf 63.6, 66.7 & 52.7, 65.7  & -\\
			MT-DrQA on Sv2+W & -,- &63.5, 66.3& -,- & 75.4 \\
			MT-DrQA on Sv2+N+W & -,- & 63.1, 66.3 & 52.5, 65.6  & 75.3\\
\hline
			SOTA (Single Model) & 80.0, 87.0 & 72.3, 74.8& 48.4, 63.7 (test) & 71.7 (test)\\
            \bf MT-DrQA Best Performance &\bf 70.2, 79.3& \bf 63.6, 66.7 & \bf 53.0, 66.2(test) & \bf 75.4 (test)\\
			Human Performance (test set) & 82.3, 91.2 & 86.8, 89.5 & 46.5, 69.4 &  84\\
			\hline
		\end{tabular}
	\end{center}
	\caption{\label{tab:mt-drqa} Single model performance of our method to train DrQA on multi-task setting, as well as state-of-the-art (SOTA) results and human performance. SQuAD and NewsQA performance are measured by (EM, F1), and WDW by accuracy percentage. All results are on development set unless otherwise noted. Published SOTA results come from \protect\cite{wang2018multib,hu2018read+,kundu2018question,yang2016words} respectively. }
\end{table*}
We apply MT-DrQA to SQuAD (v1.1 and v2.0), NewsQA, and WDW. We follow the setup of \cite{chen2017reading} for model architecture and hyperparameter setup. We use Algorithm \ref{algo:mtmrc} to train all MT-DrQA models. Different than \cite{rajpurkar2018know}, we do not optimize the evaluation score by changing the threshold to predict unanswerable question for SQuAD v2.0; we just use the argmax prediction. As a result, we expect the gap between dev and test performance to be lower for our model.
The results of MT-DrQA are presented in Table \ref{tab:mt-drqa}. The results of combining SQuAD and NewsQA obtain similar performance boost as our SAN experiment, with a performance boost between 1-2\% in both EM and F1 for the two datasets. The results of MTL including WDW is different: although adding WDW to SQuAD still brings a marginal performance boost to SQuAD, the performance on WDW drops after we add SQuAD and NewsQA into the training process. We conjecture that this negative transfer phenomenon is probably because of the drastic difference between WDW and SQuAD/NewsQA, both in their domain, answer type, and task type; and DrQA might not be capable of caputuring all these features using just one network.
We leave the problem of further preventing such negative transfer to future work.
\end{document}